\documentclass{article}

\usepackage{microtype}
\usepackage{graphicx}
\usepackage{subfigure}
\usepackage{booktabs} % for professional tables
\usepackage{caption}
\usepackage{enumitem}
\usepackage{makecell}
\usepackage{graphicx}
\usepackage{tikz}
\usepackage{pgfplots}
\usepackage{amsfonts}
\usepackage{hyperref}
\usepackage[toc, page]{appendix}
\usepackage{amsmath}
\DeclareMathOperator*{\argmax}{argmax}
\DeclareMathOperator*{\argmin}{argmin}

\usepackage{hyperref}

\usepackage[accepted]{icml2022}

\icmltitlerunning{Brick Tic-Tac-Toe: Exploring the Generalizability of AlphaZero to Novel Test Environments}

\begin{document}

\twocolumn[
\icmltitle{Brick Tic-Tac-Toe: Exploring the Generalizability of AlphaZero to Novel Test Environments}

% It is OKAY to include author information, even for blind
% submissions: the style file will automatically remove it for you
% unless you've provided the [accepted] option to the icml2019
% package.

% List of affiliations: The first argument should be a (short)
% identifier you will use later to specify author affiliations
% Academic affiliations should list Department, University, City, Region, Country
% Industry affiliations should list Company, City, Region, Country

\begin{icmlauthorlist}
\icmlauthor{John Tan Chong Min}{nus}
\icmlauthor{Mehul Motani}{nus}
\end{icmlauthorlist}

\icmlaffiliation{nus}{Department of Electrical and Computer Engineering, National University of Singapore}

\icmlcorrespondingauthor{John Tan Chong Min}{johntancm@u.nus.edu}
\icmlcorrespondingauthor{Mehul Motani}{motani@u.nus.edu}

% You may provide any keywords that you
% find helpful for describing your paper; these are used to populate
% the "keywords" metadata in the PDF but will not be shown in the document
\icmlkeywords{Machine Learning, Reinforcement Learning, AlphaZero, ICML}

\vskip 0.3in
]

\printAffiliationsAndNotice{} 

\begin{abstract}
Traditional reinforcement learning (RL) environments typically are the same for both the training and testing phases. Hence, current RL methods are largely not generalizable to a test environment which is conceptually similar but different from what the method has been trained on, which we term the \textit{novel test environment}. As an effort to push RL research towards algorithms which can generalize to novel test environments, we introduce the Brick Tic-Tac-Toe (BTTT) test bed, where the brick position in the test environment is different from that in the training environment. Using a round-robin tournament on the BTTT environment, we show that traditional RL state-search approaches such as Monte Carlo Tree Search (MCTS) and Minimax are more generalizable to novel test environments than AlphaZero is. This is surprising because AlphaZero has been shown to achieve superhuman performance in environments such as Go, Chess and Shogi, which may lead one to think that it performs well in novel test environments. Our results show that BTTT, though simple, is rich enough to explore the generalizability of AlphaZero. We find that merely increasing MCTS lookahead iterations was insufficient for AlphaZero to generalize to some novel test environments. Rather, increasing the variety of training environments helps to progressively improve generalizability across all possible starting brick configurations.

\end{abstract}

\section{Introduction}
Through the years, there has been tremendous progress in Reinforcement Learning (RL) techniques for both single-player environments and two-player perfect information environments. Progress has been fuelled by both hardware advancements as well as availability of computer simulation environments such as OpenAI Gym \cite{brockman2016openai}.

Current RL algorithms are largely trained and tested in the same environment, for instance Atari 2600, chess and Go. While superhuman performance have been achieved in these environments \cite{mnih2013playing, campbell2002deep, silver2016mastering}, they may not generalize well to a test environment which is conceptually similar but different from what the method has been trained on, which we term the \textit{novel test environment}. This limits their deployability to the real world, where the environment is not stationary and can change over time.

\textbf{Previous work on single-player games:}
Previous work on single-player games include continuous control tasks such as Cart Pole, Mountain Car as well as physics joint-based MUJOCO environments \cite{todorov2012mujoco}. A general survey of RL methods for these environments, typically policy gradient methods such as REINFORCE \cite{williams1992simple} or Trust Region Policy Optimization (TRPO) \cite{schulman2015trust} is detailed in \citet{duan2016benchmarking}. For the Atari 2600 arcade environment, Deep Q-Network (DQN) was a breakthrough for Model-Free RL, combining traditional Q-Learning with deep neural networks along with a replay buffer to achieve superhuman performance on a variety of Atari 2600 games \cite{mnih2013playing, mnih2015human}. 

\textbf{Previous work on two-player perfect information games:}
Previous work on two-player perfect information games are generally model-based methods used on board games and can be classified under three main approaches. The first approach is via lookahead search with the Minimax algorithm, alpha-beta pruning and heuristics for evaluating the value of a game state, and is applied in Samuel's checkers \cite{samuel1959some}, Shannon's chess \cite{shannon1950xxii}, and eventually Deep Blue's chess \cite{campbell2002deep}. The second approach is via Temporal Difference (TD) learning, most notably utilized in TD-Gammon \cite{tesauro1994td} for backgammon. The third approach is via Monte Carlo Tree Search (MCTS) \cite{chaslot2008monte} to prioritize searching for paths through the game tree which are higher in potential reward using an explore-exploit Upper Confidence Bounds in Trees (UCT) heuristic \cite{kocsis2006bandit}. This was utilized in AlphaGo \cite{silver2016mastering}, AlphaGo Zero \cite{silver2017mastering} for the game of Go. AlphaZero \cite{silver2017masteringalphazero} is an extension of AlphaGo Zero for generic games such as Go, Chess and Shogi. MuZero \cite{schrittwieser2019mastering} is an extension of AlphaZero for both one-player and two-player games using an internal learned model for planning. There have been numerous works improving or reimplementing AlphaZero \cite{bratko2018alphazero, moerland2018a0c, tian2019elf,  wang2019alternative, wang2019hyper, grill2020monte, ben2021train, dam2021convex}.

\begin{figure*}[t]
\centering
	\begin{minipage}[t]{0.22\textwidth}
		\includegraphics[width=\textwidth]{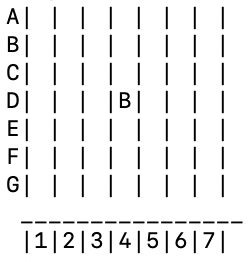}
		\caption{BTTT Environment (Variant 1) with starting brick position at D4}
		\label{fig:var1}
	\end{minipage}%
	\hfill
	\begin{minipage}[t]{0.22\textwidth}
		\includegraphics[width=\textwidth]{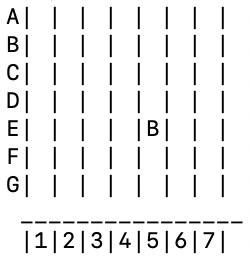}
		\caption{BTTT Environment (Variant 2) with starting brick position at E5}
		\label{fig:var2}
	\end{minipage}
	\hfill
	\begin{minipage}[t]{0.22\textwidth}
		\includegraphics[width=\textwidth]{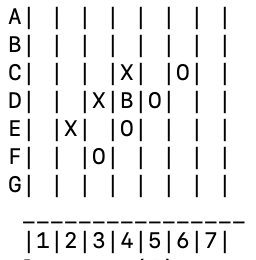}
		\caption{Player 1 (O) winning in BTTT}
		\label{fig:owin}
	\end{minipage}
	\hfill
	\begin{minipage}[t]{0.22\textwidth}
		\includegraphics[width=\textwidth]{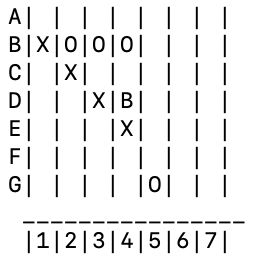}
		\caption{Player 2 (X) winning in BTTT}
		\label{fig:xwin}
	\end{minipage}
% 	\caption{add caption here}
\end{figure*}

\textbf{Limitations of current environments:} The current environments used for RL generally do not have different train and test environments. Indeed, even the pseudo-random-number-generator for Atari 2600 games were hardware-constrained by the Linear Feedback Shift Register to have a fixed random seed, causing the same deterministic environmental interactions in both train and test environments \cite{hausknecht2015impact}. Hence, while neural network-based approaches such as DQN \cite{mnih2013playing, mnih2015human} can learn without human expertise in majority of single-player Atari 2600 games, DQN may be able to just play well by memorizing winning sequences of actions. Indeed, recent papers \cite{yuan2019adversarial, qu2020minimalistic} have shown that even a one-pixel change can adversely affect outcomes in DQN, which indicates that the learned algorithm only works on an exact replication of the training environment, and is not generalizable. Likewise, although AlphaZero achieves superhuman performance for two-player perfect information games such as chess, Shogi and Go \cite{silver2017masteringalphazero}, the same phenomenon of memorizing might occur for AlphaZero when trained and tested on the same environment, which will be the focus of this paper.

% While there have been attempts to make the test environment different from training through random play before letting the agent take over, as utilized in MuZero \cite{schrittwieser2019mastering}, this approach is not ideal as the environment is fundamentally not changed - only the start state of game play is changed according to random moves. 

\textbf{Similar work:} Some progress has been made to improve quantization of generalization ability, most notably through the use of Procedural Content Generation (PCG) \cite{risi2020increasing}. One such PCG-based approach is ``Coin Run", which feature different train and test environments for a single-player game.
% Also, as a benchmark to measure general artificial intelligence, Chollet has released the Abstraction and Reasoning Corpus (ARC), which attempts to measure generalizability for various tasks for a single person by demonstrating a few training inputs and outputs, while testing on unseen inputs and outputs which require the same conceptual understanding \cite{chollet2019measure}. 
Our work is presented in a similar vein to that of PCG-based approaches - to quantize and develop more generalizable deep RL algorithms in a two-player setting. One similar work exploring generalizability in two-player games is \citet{ben2021train}, which has focused on generalizability to different board sizes via infusing AlphaZero with a graph neural network (GNN). Another work focuses on zero-shot transfer learning of AlphaZero-like network weights between game variants and varying board sizes \cite{soemers2021transfer}. Another work similar to our objective is XLand, an open-ended environment which was recently created to test generalization to novel environments. However, XLand is very complex and requires billions of timesteps in order to train an agent \cite{openendedlearningteam2021openended}, which limits fast iteration of algorithms. Our proposed environment similarly is able to test generalization to novel environments, but only requires hundreds of iterations, each using thousands of timesteps, to train the agent. In this paper, we demonstrate that our proposed environment is sufficient to learn about the generalizability of AlphaZero. Furthermore, our environment has an advantage that we can use heuristics to create an agent which achieves perfect play using Minimax, which helps facilitate benchmarking our other agents.

\textbf{Motivation:} In order to test students' understanding, a good teacher would set an examination which is structurally different from the practice problems, yet conceptually similar. Similarly, in order to test the generalizability of current RL algorithms such as AlphaZero, MCTS, Minimax, we should use a slightly different test environment to see if they merely memorize the ``model answer" or are able to learn the underlying concepts. To this end, we propose a test bed called Brick Tic-Tac-Toe (BTTT) where the test environment is different from the training environment, so as to better model changing environmental conditions in real-life. In BTTT, the first player can theoretically always win under perfect play using the same learned strategy in the training environment.

\textbf{Our Contributions:} The contributions of this paper are:
\begin{itemize}[leftmargin=*,topsep=0pt,noitemsep]
\item In order to analyze the generalizability of current RL algorithms for two-player games, we introduce our own test bed, BTTT. The BTTT environment code and code for the experiments can be found \href{https://github.com/tanchongmin/BTTT}{here}.

\item We pit various agents using different RL methods in a round-robin tournament on the BTTT environment. We observe that state-search approaches such as MCTS and Minimax are more generalizable than AlphaZero. While all algorithms can win as first player in the training environment, only AlphaZero fails to win as first player in a novel test environment.

\item We investigate the effect of changing the parameters of AlphaZero on its generalizability. We find that increasing MCTS lookahead iterations was insufficient for AlphaZero to generalize to some novel test environments. Rather, increasing the variety of training environments helps to progressively improve generalizability across all possible starting brick configurations. Specifically, we show that it is more generalizable after training on at least two distinct starting brick positions, and fails to generalize when there is only one brick starting position.

\end{itemize}

\section{Brick Tic-Tac-Toe environment}

Current popular learning environments (e.g., Cart Pole, Mountain Car, MUJOCO, checkers, chess, Go, Atari 2600) typically have the same environment for both training and testing and are not a good proxy to our real-world environment where the environment itself can change over time. 

To enable the study of RL methods under changing environments, we propose to use a test bed, \textbf{Brick Tic-Tac-Toe (BTTT)}, to serve as a controlled environment to evaluate the performance of various RL methods under a conceptually similar, yet unseen test environment. While BTTT might be a simple environment, it allows us to do a comprehensive analysis of generalizability across various algorithms, which makes it a suitable testbed for quick benchmarking of generalizability. 

The BTTT environment is a two-player perfect information, sequential turn-based environment. The following details the core components of the BTTT environment:

\textbf{State/Board:} The game is played on a 7x7 grid, with the rows labelled A to G, and columns labelled 1 to 7. Each grid square has a unique name derived from the row then column, e.g. A3, E5.

\textbf{Pieces:} There are three pieces, O for Player 1's piece, X for Player 2's piece, and B for a brick block. If there are no pieces on a grid square, it is represented as a blank space.

\textbf{Start State:} At start state, a grid square will be chosen to be a brick block B, making that square unplayable by both players. The remaining squares will be empty. The start state will be known to both players.

\textbf{Rules:} The game is a two-player game played sequentially on a 7x7 grid. It is similar in concept to Tic-Tac-Toe and Gomoku \cite{tang2016adp}, except that the aim of the game is to form a 4-in-a-row, for both players 1 (O) and 2 (X). O plays first, followed by X, and so on, until there is a winner. If the game is not won and all possible squares are filled up, it is counted as Player 2's (X) win. 

\textbf{Training Environment (Variant 1):} This is a 7 x 7 grid with a brick block B in the centre, position D4. This is depicted in Fig. \ref{fig:var1}.

\textbf{Test Environment (Variant 2):} This is a 7 x 7 grid with a brick block B at position E5. This is depicted in Fig. \ref{fig:var2}.

\textbf{Win Condition:} A win occurs after a player's move when there is a 4-in-a-row of Os or Xs in any row, column or diagonal. Note that there is no wraparound in BTTT, that is, the left and right side of the grid, or the top and bottom of the grid do not connect. An example of Player 1's win (O) is depicted in Fig. \ref{fig:owin}, and an example of Player 2's win (X) is depicted in Fig. \ref{fig:xwin}. 

\textbf{Perfect Play:} BTTT is designed for the first player to win (Refer to \textbf{Appendix \ref{section:winnable}}), which means that an agent that is generalizable should be able to win as first player regardless of where the brick block appears at start state. 

% \section{Stochastic Concept-Aware Model}

% Still thinking about this, possibly has to do with online learning in new environment, and also to do with some form of self-modelling of the world.

% Refer to \textbf{Algorithm \ref{alg:Stochastic}} for the psuedo-code.

% \begin{algorithm}[!t]
% 	\caption{Stochastic Concept-Aware Model}
% 	\label{alg:Stochastic}
% 	\begin{algorithmic}
% 		\STATE {\bfseries Input:} xx
% 		\STATE {\bfseries Hyperparameters:} xx
% 		\REPEAT
% 		\STATE 1. xx
% 		\STATE 2. xx
% 		\STATE 3. xx
% 		\STATE 4. xx

% 		\UNTIL xx
% 	\end{algorithmic}
% \end{algorithm}

\section{Algorithm Details}
We firstly go through some algorithm details for the agents which we use for the tournament.

\subsection{Markov Decision Process}
We will model the BTTT environment as a Markov Decision Process (MDP), whereby for a policy mapping a state to an action distribution, $\pi: S \to A$, only the current state $s_t$ is needed as input and the policy is independent of all previous states. That is, 
\begin{equation}
    \label{eqn:MDP}
    \pi(s_t|s_1, s_2, s_3, ..., s_{t-1}) = \pi(s_t),
\end{equation}
where $s_i$ represents the state of the game after action $i$. 

From (\ref{eqn:MDP}) above, we can see that given the current state of the game $s_t$, an optimal player has all the information he/she needs to decide on the next action/move (action space is all of the empty grid squares). Unlike other games like Chess and Shogi \cite{silver2017masteringalphazero}, there is no need to maintain a history of past states in the MDP formulation when doing action-selection, since there will not be a case of repeated game states in the same game.

The formulation of the environment as a MDP helps facilitate applying tree-search based methods like Minimax and MCTS. At each node in the tree, we can choose our next action based on just on the game state of that node itself.

% \begin{figure*}[t]
% \centering
% 	\begin{minipage}[t]{\textwidth}
% 		\includegraphics[width=\textwidth]{Pictures/Monte Carlo Tree Search.png}
% 		\caption{Monte Carlo Tree Search procedure. Image taken from \cite{chaslot2008monte}.}
% 		\label{fig:montecarlo}
% 	\end{minipage}%
% % 	\caption{add caption here}
% \end{figure*}

\begin{table*}[h]
\begin{center}
\begin{tabular}{|c|p{12cm}|}
\hline
\textbf{Parameter} & \textbf{Value}\\
\hline
Optimizer & Stochastic Gradient Descent (SGD), learning rate 0.1, momentum 0.9\\
Exploration factor $c_{PUCT}$ & 1\\
Memory size per iteration & 6000\\
Batch size for training & 256\\
MCTS simulations per move & 100\\
Regularization constant $c$ & 0.0001\\
\hline
Architecture (\textbf{Input}) & Input layer of three binary planes of 7x7, followed by a Conv2D (75) block\\
Architecture (\textbf{Body}) & From \textbf{Input}, ResNet architecture with 5 hidden layers of two Conv2D (75) blocks each\\
Architecture (\textbf{Output - Value}) & From \textbf{Body}, Conv2D (1) block, a Flatten layer, a Dense layer with 20 nodes, LeakyReLU activation function, a Dense layer with 1 node\\
Architecture (\textbf{Output - Policy}) & From \textbf{Body}, Conv2D (2) block, a Flatten layer, a Dense layer with 49 nodes\\

\hline
\end{tabular}
\caption{Architecture and Hyperparameters of AlphaZero. Each Conv2D (X) block has this structure: Conv2D with X filters of 4x4, 'same' padding, followed by BatchNormalization, followed by the LeakyReLU activation function. For the Conv2D (X) blocks at the output, the filter size is 1x1 instead.}
\label{table:AlphaZero architecture}
\end{center}
\end{table*}

\subsection{Minimax}

Minimax has been used successfully in world champion chess programs such as Deep Blue \cite{campbell2002deep}, and is a tree-based search method which models both players acting sequentially in an optimal manner. In particular, we focus on a two-ply lookahead Minimax which considers all board states $s_{t+2}$ two moves ahead. In particular, for board state $s_t$ which contains all actions till action $t-1$: $a_1, a_2, ..., a_{t-1}$, if it is the maximizing player's move, we choose the next action $a_{t}$ for the maximizing player as

\begin{equation}
\label{eqn:maxplayer}
a_{t} = \argmax_{a_{t}}\min_{a_{t+1}}V(s_{t+2}|s_t, a_{t}, a_{t+1}).
\end{equation}

Similarly, if it is the minimizing player's move, we choose the next action $a_{t}$ for the minimizing player as

\begin{equation}
\label{eqn:minplayer}
a_{t} = \argmin_{a_{t}}\max_{a_{t+1}}V(s_{t+2}|s_t, a_{t}, a_{t+1}).
\end{equation}

In (\ref{eqn:maxplayer}) and (\ref{eqn:minplayer}) above, $V(.)$ represents a heuristic-based value function that evaluates the game state. A positive value represents a state which is good for the maximizing player, while a negative value represents a state which is good for the minimizing player. The higher the magnitude, the more desirable the state for that player.

\subsection{MCTS}

MCTS, like Minimax, is a tree-based search method, but it allows the game tree to be searched efficiently by prioritizing the moves which lead to higher rewards in an explore-exploit manner. It comprises of four steps at each iteration.
% as shown in Fig. \ref{fig:montecarlo}.

\textbf{Selection.} At each node (state) in the game tree, the next action is chosen in a way to balance between exploration and exploitation. This is done using the Upper Confidence Bound applied to Trees (UCT) algorithm \cite{kocsis2006bandit}.

We choose the child node by the UCB1 formula
\begin{equation}
\label{eqn:UCB}
    \argmax_{a}{\frac{v_a}{n_a} + C\times\sqrt{\frac{\ln(\sum_{b \in A}{n_b})}{n_a}}},
\end{equation}

where $C$ is the exploration constant, which we set as 1, $v_a$ is the value of the child node taking action $a$ based on the total value of all nodes below it, $n_a$ is the number of visits of the child node taking action $a$, $A$ represents the entire set of valid actions from the parent node.

\textbf{Expansion.} Once the tree reaches a leaf node which is not fully expanded (all child nodes created), we expand out a new node by selecting an untried action from this leaf node.

\textbf{Simulation.} The outcome of a game from the leaf node is simulated by a random rollout, that is, by playing random moves from the game state of the node to the end of the game. Heuristic-based action-selection could be used to make the rollout more accurate, but is not used in this paper.

\textbf{Backpropagation.} At the end of the game, the value of each node traversed during the game is updated with the game result $r$, which is 1 for Player 1's win, -1 for Player 2's win, and 0 for a draw. For a node below a maximizing player, $v_a = v_a + r$, while for a node below a minimizing player, $v_a = v_a - r$. This ensures that the value of the game result is converted to a form where both players can maximize to choose the optimal action based on (\ref{eqn:UCB}). Also, for each node traversed during the game, the visit counts are increased by one: $n_a = n_a + 1$.

Finally, after going through the iterations, the action selected is simply the child node which was explored the most, which should lead to the highest reward.

\subsection{AlphaZero}

AlphaZero combines neural networks (i.e. policy and value networks) with MCTS. The policy network helps to narrow down the search at each game state $s_t$ to moves which can lead to good outcomes, while the value network helps to give a value for the game state as an approximation for lookahead search. This was first utilized in AlphaGo \cite{silver2016mastering}, and then AlphaGo Zero \cite{silver2017mastering} for the game of Go. We use the methodology of AlphaZero, that is, \textit{tabula-rasa} without human expert knowledge \cite{silver2017mastering}. For the self-play iterations, we update the best agent after each self-play iteration, without the need for tournament evaluation, just like in \citet{silver2017masteringalphazero}.

The neural network in AlphaZero starts off with randomly initialized parameters $\theta$. The self-play games are played by selecting actions via MCTS for both players, $a_t \sim \pi_t$. The terminal position $s_T$ is scored as $z$: -1 for a loss, 0 for a draw and +1 for a win. At the end of a batch of self-play games, the network parameters $\theta$ are updated to minimize the error of the predicted outcome $v_t$ and the actual game outcome $z$ using mean-squared error, as well as maximize the similarity of the policy vector $p_t$ to MCTS search probabilities $\pi_t$ using Kullback-Leibler (KL) divergence, with an L2 weight regularization term parameterized by $c$.

The loss function used to update the neural network is

\begin{equation}
    l = (z-v)^2 - \pi^T\log p + c{||\theta||}^2,
\end{equation}

with $(s,\pi,z)$ sampled uniformly across all time steps $(s_t,\pi_t,z_t)$ of self-play games in the last agent iteration, while the policy and value network values are derived from
\begin{equation}
    (p,v) = f_\theta(s).
\end{equation}

The architecture of the neural network used follows closely to that of \citet{silver2017mastering}, and is detailed in Table  \ref{table:AlphaZero architecture}. The feature planes used are a 7x7 binary plane for the current players' pieces, a 7x7 binary plane for the next players' pieces, and a 7x7 binary plane for brick pieces. This differs slightly from \citet{silver2017mastering} as we do not have a plane to encode player turn, but rather, we always let the first 7x7 binary plane be that of the current player so that it will always be invariant to either player. 

Similar to the AlphaZero papers \cite{silver2017masteringalphazero,silver2017mastering}, in order to enable sufficient exploration in beginning states, we pick a move from the move probability distribution of the visited moves of the MCTS searches in the first 10 turns, before reverting to greedy action selection based on the most visited move from the MCTS search. Also, dirichlet noise was added during training (removed during testing) to bias move selection from the root node in MCTS so as to increase move diversity.

\begin{table*}[t]
\begin{center}
\begin{tabular}{|c|c|c|c|}
\hline
\textbf{4-in-a-row Configurations} & \textbf{Value}\\
\hline
X\_\_\_ or \_\_\_X & -0.000001\\
\_X\_\_ or \_\_X\_ & -0.000002\\
X\_X\_ or X\_\_X or \_X\_X or XX\_\_ or \_\_XX & -0.0001\\
\_XX\_ & -0.0002\\
X\_XX or XX\_X or XXX\_ or \_XXX & -0.01\\
XXXX & -1\\
\hline
O\_\_\_ or \_\_\_O & 0.0000015 \\
\_O\_\_ or \_\_O\_ & 0.000003 \\
O\_O\_ or O\_\_O or \_O\_O or OO\_\_ or \_\_OO & 0.00015\\
\_OO\_ & 0.0003\\
O\_OO or OO\_O or OOO\_ or \_OOO & 0.015\\
OOOO & 1.5\\
\hline
Other & 0\\
\hline
\end{tabular}
\caption{Heuristic used to evaluate state position of the 7x7 grid in Minimax. `\_' represents a blank grid square. Value of a grid state is the sum of values of all possible 4-in-a-row configuration values across all rows, columns, diagonals in the 7x7 grid. Each 4-in-a-row configuration value is determined by reading off this lookup table. Higher value is given to more pieces of the same kind, and configurations with blanks on both sides of them get higher weightage due to increased possibilities of an extension later in the game. The 4-in-a-row configuration for Player 1 (O) gets 1.5 times the weightage of the corresponding ones for Player 2 (X), so as to encourage Player 2 to block Player 1, as Player 1 has the initiative and needs to be prevented from winning before Player 2 can win.}
\label{table:minimaxheuristic}
\end{center}
\end{table*}

\section{Methodology}
In this section, we describe the BTTT tournament in which various agents are pit against each other.

\subsection{Tournament Format}
We run a round-robin tournament of 100 rounds between various agents using Variant 2 in order to determine the most generalizable agent. At each turn, the input to the agent will be the board state of each of the 7x7 grid squares, and the output will be the action index indicating the selected square to place the piece on. We then evaluate an agent's performance by the number of wins it obtained over 100 games: the higher the number, the better the performance. 

\subsection{Agents used in the Tournament}

\begin{itemize}[leftmargin=*,topsep=0pt,noitemsep]
\item \textbf{Random.} This agent simply selects a random (valid) action each turn. This is the weakest baseline and all other methods are supposed to be stronger than this.

\item \textbf{MCTS 1000.} This is MCTS with 1000 iterations. This serves as a weak baseline to evaluate other models.

\item \textbf{MCTS 10000.} This is MCTS with 10000 iterations. This serves as a strong baseline to evaluate other models.

\item \textbf{Minimax.} This is applying the Minimax tree search with alpha-beta pruning to select optimal actions by expanding the game tree to depth 2, and thereafter using an expert heuristic (see Table \ref{table:minimaxheuristic}) to determine the value of a game state. This serves as the strongest baseline to evaluate other models, and is a proxy to perfect play.

\item \textbf{AlphaZero 100.} An AlphaZero agent was trained from self-play using MCTS lookahead search to perform policy improvement and evaluation as suggested by the AlphaZero papers \cite{silver2017masteringalphazero,silver2017mastering}. The architecture is similar to the generic version of AlphaZero \cite{silver2017masteringalphazero}, and we augment the memory with all 4 reflections of the game due to the symmetrical nature of the game, which helps the neural network to train faster. We use 100 MCTS lookahead steps for each move. Each iteration of AlphaZero trains until it has 6000 new states in memory (including reflections) before using minibatch sampling to train the model for the next iteration. We train it for two weeks, for a total of 850 iterations.

\item \textbf{AlphaZero 1000.} This is the same as AlphaZero 100 during training, except that during testing, we use 1000 MCTS lookahead searches.

\item \textbf{AlphaZero NS.} This is the same as AlphaZero 100 during training, but with no MCTS lookahead search during testing. The agent simply selects the move with the highest move probability based on the policy network. This agent is used as a control to evaluate the role of the MCTS lookahead in the generalizability of the network.

\end{itemize}

\subsection{Training Procedure}

There is no training required for MCTS and Minimax. For AlphaZero, which utilizes a deep learning model, we train the model under Variant 1, and test it under both variants. We train AlphaZero for up to two weeks (850 iterations). If the model has sufficiently learned the concept of winning with a brick under Variant 1 and not just memorize the sequence of winning moves in Variant 1, it should win all the time under Variant 2 as well, since the game is designed to be always winnable by the first player under perfect play.

\begin{table}[t]
\begin{center}
\begin{tabular}{|c|c|c|c|}
\hline
\textbf{Player 1} & \textbf{Player 2} & \textbf{Result} & \textbf{Result}\\
&& (Var 1) & (Var 2)\\
\hline
Any & Random & 100 - 0 & 100 - 0\\
Random & Any & 0 - 100 & 0 - 100\\
Random & Random & 60 - 40 & 53 - 47\\
\hline
\end{tabular}
\caption{Results of Random agent versus Any (AlphaZero NS/100/1000, MCTS 1000/10000, Minimax) and also Random agent against itself in a round-robin tournament of 100 rounds, under both Variant 1 and Variant 2 environments. AlphaZero is trained only on Variant 1, but tested on both variants.}
\label{table:randomresults}
\end{center}
\end{table}

\begin{table}[t]
\begin{center}
\begin{tabular}{|c|c|c|c|}
\hline
\textbf{Player 1} & \textbf{Player 2} & \textbf{Result} & \textbf{Result}\\
&& (Var 1) & (Var 2)\\
\hline
MCTS 1000 & MCTS 1000 & 70 - 30 & 70 - 30\\
MCTS 1000 & MCTS 10000 & 15 - 85 & 15 - 85\\
MCTS 10000 & MCTS 1000 & 100 - 0 & 100 - 0\\
MCTS 10000 & MCTS 10000 & 70 - 30 & 70 - 30\\
Minimax & MCTS 1000 & 100 - 0 & 100 - 0\\
Minimax & MCTS 10000 & 100 - 0 & 100 - 0\\
Minimax & Minimax & 100 - 0 & 100 - 0\\
MCTS 1000 & Minimax & 9 - 91 & 9 - 91\\
MCTS 10000 & Minimax & 51 - 49 & 51 - 49\\
AlphaZero NS & Minimax & 100 - 0 & 0 - 100\\
AlphaZero 100 & Minimax & 100 - 0 & 0 - 100\\
AlphaZero 1000 & Minimax & 100 - 0 & 0 - 100\\
Minimax & AlphaZero 1000 & 100 - 0 & 100 - 0\\
\hline
\end{tabular}
\caption{Results of round-robin tournament of 100 rounds. AlphaZero is trained only on Variant 1, but tested on both variants.}
\label{table:results}
\end{center}
\end{table}

\begin{table}[t]
\begin{center}
\begin{tabular}{|c|c|}
\hline
\textbf{Agent} & \textbf{Time per move (s)}\\
\hline
Random & 0.000006\\
MCTS 1000 & 1.88\\
MCTS 10000 & 18.8\\
Minimax & 0.97\\
AlphaZero NS & 0.51\\
AlphaZero 100 & 5.99\\
AlphaZero 1000 & 54.87\\
\hline
\end{tabular}
\caption{Time per move for each agent, averaged over 100 moves.}
\label{table:timetaken}
\end{center}
\end{table}

\subsection{Runtime Analysis Procedure}

In order to measure the tradeoff between playing strength and runtime, we perform the runtime analysis by calculating the average time taken per move over 100 moves for the agent to choose a move/action given the initial board state. The runtime was calculated using an NVIDIA 2080 RTX GPU and a 256GB RAM CPU.

\section{Tournament Results}

\begin{table*}[t]
\centering
\begin{tabular}{|c|c|c|c|c|}
\hline
\textbf{Player 1} & \textbf{Player 2} & \textbf{Result (Var 1)} & \textbf{Result (Var 2)}\\
% \hline
% AlphaZero NS (trained under D4 (Var 1) only) \textbf{[Baseline]} & Minimax & 100 - 0  & 0 - 100\\
% AlphaZero NSR2 (trained randomly under D4 (Var 1), D3) & Minimax & 0 - 100 & 0 - 100\\
% AlphaZero NSR3 (trained randomly under D4 (Var 1), D3, C3) & Minimax & 0 - 100 & \textbf{84 - 16}\\
\hline
AlphaZero 100 (trained under D4 only) \textbf{[Baseline]} & Minimax & 100 - 0  & 0 - 100\\
AlphaZero 100R2 (trained randomly under D3 and D4) & Minimax & 100 - 0 & \textbf{100 - 0}\\
AlphaZero 100R3 (trained randomly under C3, D3 and D4) & Minimax & 100 - 0 & \textbf{100 - 0}\\
\hline
AlphaZero 1000 (trained under D4 (Var 1) only) & Minimax & 100 - 0  & 0 - 100\\
\hline
\end{tabular}
\caption{Additional results of AlphaZero vs Minimax in 100-round games, under both Variant 1 and Variant 2 environments.}
\label{table:alphazerovariantresults}
\end{table*}

\begin{table*}[t]
\centering
\begin{tabular}{|c|c|c|c|c|}
\hline
\textbf{Player 1} & \textbf{Player 2} & \textbf{Result} & \textbf{Player 1 Win Rate(\%)}\\
% \hline
% AlphaZero NS (trained under D4 (Var 1) only) \textbf{[Baseline]} & Minimax & 44 - 446\\
% AlphaZero NSR2 (trained randomly under D3 and D4) & Minimax & 95 - 395\\
% AlphaZero NSR3 (trained randomly under C3, D3 and D4) & Minimax & 463 - 27\\
\hline
AlphaZero 100 (trained under D4 only) \textbf{[Baseline]} & Minimax & 139 - 351 & 28.4\\
AlphaZero 100R2 (trained randomly under D3 and D4) & Minimax & 306 - 184 & 62.4\\
AlphaZero 100R3 (trained randomly under C3, D3 and D4) & Minimax & 443 - 47 & 90.4\\
\hline
AlphaZero 1000 (trained under D4 only) & Minimax & 364 - 126 & 74.3\\
AlphaZero 1000R2 (trained randomly under D3 and D4) & Minimax & 415 - 75 & 84.7\\
AlphaZero 1000R3 (trained randomly under C3, D3 and D4) & Minimax & 483 - 7 & 98.6\\
\hline
Minimax \textbf{[Benchmark for Generalizable Perfect Play]} & Minimax & 490 - 0 & 100\\
\hline
\end{tabular}
\caption{Additional results of AlphaZero vs Minimax in 490-round games, with 10 games for each starting brick position.}
\label{table:alphazerovariant2results}
\end{table*}

%9293 - R3, 9294 - Baseline, 9295 - R2

%1000 iterations results
%Baseline: 364-126
%R2: 415-75
%R3: 483-7

%Current status
%9293 - R3, 9294 - R2, 9295 - Computing 1000 iterations

\textbf{Summary:} The results of the round-robin tournament are detailed in Tables \ref{table:randomresults} and \ref{table:results}. We observe that Minimax achieves perfect play, while AlphaZero trained on only Variant 1 fails to generalize to Variant 2.

\textbf{Random Agent vs MCTS/Minimax/AlphaZero:} From Table \ref{table:randomresults}, we can see that the random agent loses all games against MCTS/Minimax/AlphaZero, regardless whether it is Player 1 or Player 2, under both Variants 1 and 2. We conclude that the MCTS, Minimax and AlphaZero agents perform better than random under any general environment.

\textbf{MCTS vs Minimax:} From Table \ref{table:results}, we can conclude that MCTS 1000 and MCTS 10000 are not playing perfectly as they lose some games as Player 1 when pit against Minimax as Player 2. This is likely because Minimax has expert knowledge integrated into the heuristic used to evaluate the value function of the game state, and as such performs the best.  MCTS evaluates the game state by random rollouts, and just 10000 iterations worth of rollouts is probably insufficient to give a proper valuation of the game state due to the large game state space.

\textbf{Minimax vs MCTS:} From Table \ref{table:results}, we can conclude that Minimax is superior to MCTS as it claims victory in all games as Player 1 and MCTS as Player 2. Minimax typically employs a strategy similar to that of perfect play.

\textbf{Minimax/MCTS self-play:} From Table \ref{table:results}, when Minimax is pit against itself, it wins all 100 games as Player 1. This shows that Minimax is playing perfectly as it wins all the time as Player 1. The same cannot be said for MCTS, as even MCTS 10000 only wins 70 out of 100 of its games as Player 1, which shows that the state space of BTTT is so large that 10000 lookahead steps cannot guarantee perfect play. That said, MCTS 10000 is clearly superior to MCTS 1000, as the former wins all 100 of its games against the latter regardless as Player 1, but the converse is not true.

\textbf{AlphaZero vs Minimax:} From Table \ref{table:results}, we can see that AlphaZero as Player 1 is able to win all games in Variant 1 against Minimax, but is unable to do so in a slightly different Variant 2. This hints that AlphaZero may be overfitting to the training environment and hence unable to generalize to novel environments. Surprisingly, AlphaZero NS also managed to win all games in Variant 1, which means that the neural network has already achieved perfect play in the training environment, even without any lookahead search. This validates the power of the self-play algorithm of AlphaZero to improve itself as even the policy network alone is sufficient to make a good decision after enough training. 

\textbf{Minimax vs AlphaZero:} As a control, we pit Minimax as Player 1 against AlphaZero as Player 2, and Minimax managed to win all games in both variants. This means that a truly generalizable method such as Minimax should be able to do so, and hence AlphaZero in its native form is not generalizable enough.

\section{Runtime Analysis}

From Table \ref{table:timetaken}, we can see that the agents which are more generalizable (MCTS, Minimax) typically are those with lookahead search, and take more time to evaluate the position the more lookahead is done. This can be seen as MCTS 10000 has a lookahead of 10000 steps, and is about ten times the runtime of MCTS 1000, which has a lookahead of 1000 steps. Minimax looks ahead for two ply, which is $48 \times 47 = 2256$ moves. We would expect Minimax to have a runtime in between that of MCTS 1000 and MCTS 10000. However, because Minimax uses a heuristic to evaluate a leaf node (no random rollouts needed like MCTS) and utilizes alpha-beta pruning, it is actually much faster and is about half the runtime of MCTS 1000. This shows that having a heuristic can help reduce runtime significantly.

For a neural network-based value approach like AlphaZero, it is slower than the lookahead search methods, since it executes both lookahead and running through the neural network algorithm. However, AlphaZero NS with no lookahead performs significantly faster than the lookahead methods.

Overall, there is a tradeoff between playing strength and runtime. Rather than increasing playing strength by increasing lookahead steps and incurring an increase in runtime, feasible alternatives are to use a heuristic (Minimax) or to use a neural network (AlphaZero NS).

\section{Improving AlphaZero}

In Table \ref{table:results}, we can see that AlphaZero does not generalize well. We investigate if we can make AlphaZero generalize better by training with an increased variety of starting brick positions. We train two other variants of AlphaZero 100, namely AlphaZero 100R2 (starting brick position randomly at D3 and D4), and AlphaZero 100R3 (starting brick position randomly at C3, D3 and D4), In Table \ref{table:alphazerovariantresults}, we compare the performances of AlphaZero 100, 100R2 and 100R3 against the strongest baseline for perfect play (Minimax) in both Variants 1 and 2.

\textbf{Results:} From Table \ref{table:alphazerovariantresults}, AlphaZero 100R2 and AlphaZero 100R3 perform better than the baseline, achieving perfect play even in Variant 2. We posit that the varying starting brick positions could have enabled the network to learn the function of the brick block, rather than just memorizing it as a fixed position. It may hint that there may have been overfitting occuring in the AlphaZero baseline when it was only exposed to Variant 1, as even with 1000 MCTS lookahead searches, it was unable to generalize to Variant 2.

\section{Can we generalize even more?}

Given the similar performance of AlphaZero 100R2 and 100R3 in Table \ref{table:alphazerovariantresults}, we perform more experiments to further distinguish their generalizability. Instead of just evaluating on Variant 2, we evaluate on all possible starting brick positions, for a total of 490 games, with 10 games for each possible starting brick position. In Table \ref{table:alphazerovariant2results}, we compare the performance of AlphaZero 100, 100R2 and 100R3 and the variants with 1000 MCTS lookahead searches during testing (AlphaZero 1000, Alphazero 1000R2, AlphaZero 1000R3).

\textbf{Results:} The AlphaZero 100 baseline does very poorly and is only able to win in a small number of games (139 out of 490). This is likely due to the fact that the model did not manage to have sufficient examples to learn the function of a brick block. AlphaZero 100R2 performed better, winning 306 out of 490 games. This shows that even just two different starting brick positions is able to help achieve a higher level of generalizability. AlphaZero 100R3 performed even better, winning 443 out of 490 games. While AlphaZero 100R3 achieves a higher number of wins as compared to AlphaZero 100 baseline, it does not win all games and we posit that we would need more training environments in order to achieve generalizable perfect play. 

The AlphaZero 1000 variants perform better than their AlphaZero 100 counterparts, highlighting the usefulness of increased lookahead searches in generalizing to most environments (though not for all as seen in Table \ref{table:alphazerovariantresults}). 

\textbf{Towards Perfect Play:} Even with 1000 MCTS lookahead searches and three training environments, AlphaZero 1000R3 still did not manage to win all games, which highlights the complexity of the BTTT environment. As a benchmark, Minimax wins all games regardless of brick position as first player, which will be something to strive towards for neural network-based methods such as AlphaZero.

% \section{Quantizable Metric for Generalizability}

% Given that Minimax is almost at perfect play, and is about twice as fast as MCTS 1000 while achieving much stronger playing strength, we recommend the following procedure to quantize the generalizability of an agent in the BTTT environment:

% \begin{enumerate}
%     \item Train the environment under Variant 1, or any other starting position other than Variant 2 (brick at E5).
%     \item Test the environment under Variant 2 with the agent as Player 1 and Minimax as Player 2.
%     \item The generalizability score, which we term as \textit{BTTT score}, will be the number of games out of 100 won against Minimax under Variant 2.
% \end{enumerate}

% Using this metric for generalizability, the \textit{BTTT score} for AlphaZero will be 0/100 when trained under Variant 1, and 100/100 when trained randomly under Variant 1, D3, C3. MCTS 1000 will be 9/100, MCTS 10000 will be 51/100, Minimax will be 100/100.

\section{Conclusion}

BTTT is a novel testbed to study generalizability of RL methods and we use it to study the performance of AlphaZero.
Our results show that Minimax performs the best for BTTT, followed by MCTS, AlphaZero, then random. When trained solely on one training environment, AlphaZero seems to memorize the training environment and is only effective within the same training environment. It is unable to extrapolate well to novel unseen test environments, which limits its deployment to the real world. 
Our results also show that AlphaZero generalizes better to novel test environments when trained with more diverse environments, as well as with increased lookahead searches.

Overall, we can see that although BTTT is a simple environment, it allows us to do a comprehensive analysis of generalizability across various algorithms, which makes it a suitable testbed for quick benchmarking of generalizability.

\newpage
{\bf Future Work:} Given that AlphaZero is able to generalize better with increased variability in the training environments, it is interesting to see whether MuZero \cite{schrittwieser2019mastering} is able to generalize with even lesser variability of the training environments, as it internally keeps a learned model of the environment, which may help it plan better. Furthermore, given the success of GNN-based methods to generalize to different board sizes \cite{ben2021train}, we would like to explore if using GNN as the internal architecture to encompass the state information in AlphaZero/MuZero could help it generalize better as well.
Furthermore, we can also scale BTTT to larger grids with varying number of bricks to test out the effects of more complex environments on generalizability.
\newpage
% {\bf \large Acknowledgements}

% \textcolor{black}{
% This research is supported by the National Research Foundation, Singapore under its AI Singapore Programme (AISG Award No: AISG-GC-2019-002). Any opinions, findings and conclusions or recommendations expressed in this material are those of the author(s) and do not reflect the views of National Research Foundation, Singapore.}

% Acknowledgements should only appear in the accepted version.
%\section*{Acknowledgements}
%
%\textbf{Do not} include acknowledgements in the initial version of
%the paper submitted for blind review.
%
%If a paper is accepted, the final camera-ready version can (and
%probably should) include acknowledgements. In this case, please
%place such acknowledgements in an unnumbered section at the
%end of the paper. Typically, this will include thanks to reviewers
%who gave useful comments, to colleagues who contributed to the ideas,
%and to funding agencies and corporate sponsors that provided financial
%support.

% In the unusual situation where you want a paper to appear in the
% references without citing it in the main text, use \nocite
%\newpage
\bibliography{icml2022.bib}
\bibliographystyle{icml2022}

% \section{Supplementary Material}
% In this supplementary material, we present the following:
% \begin{enumerate}
%     \item Proof that Brick Tic-Tac-Toe (BTTT) is always winnable as Player 1 regardless of brick starting location.
%     \item Detailed description of the state, action, reward used in DQN for BTTT, as well as further experiments performed to enhance DQN's generalizability.
%     \item Analysis of games played between Minimax, Monte Carlo Tree Search (MCTS) and Deep Q-Network (DQN).
% \end{enumerate}

\newpage
\onecolumn
\setcounter{table}{0}
\renewcommand{\thetable}{\Alph{section}\arabic{table}}
\setcounter{figure}{0}
\renewcommand{\thefigure}{\Alph{section}\arabic{figure}}
\appendix
\section{Proof that BTTT is always winnable as first player}
\label{section:winnable}

This section demonstrates why BTTT is always winnable as first player, for Variant 1, Variant 2 as well as for any arbitrary starting brick location.

\subsection{Brick at D3 location (Variant 1)}

\textbf{Key Concept.} The key concept behind BTTT is to form multiple threat axes with the pieces. Each piece can form a 4-in-arow in four different orientations - either row, column, forward diagonal or backward diagonal, making there exist four threat axes. The moment one player has a 3-in-a-row with the left and right flank empty (we term this the \textbf{Unblockable-3}), then there is nothing the other player can do to stop him/her from forming 4-in-a-row as blocking either the left or right flank would still allow that player to put his/her piece on the other flank to form the 4-in-a-row.

\textbf{Winning Strategy Idea.} The idea behind the sure-win strategy for Player 1 is to place pieces to maximize threat axes, and after three moves, have two axes which could be expanded to form an \textbf{Unblockable-3}. The next step will then be to form the \textbf{Unblockable-3} regardless of which axis the opponent blocks and win the game.

\begin{figure*}[h]
\centering
	\begin{minipage}[h]{\textwidth}
		\includegraphics[width=\textwidth]{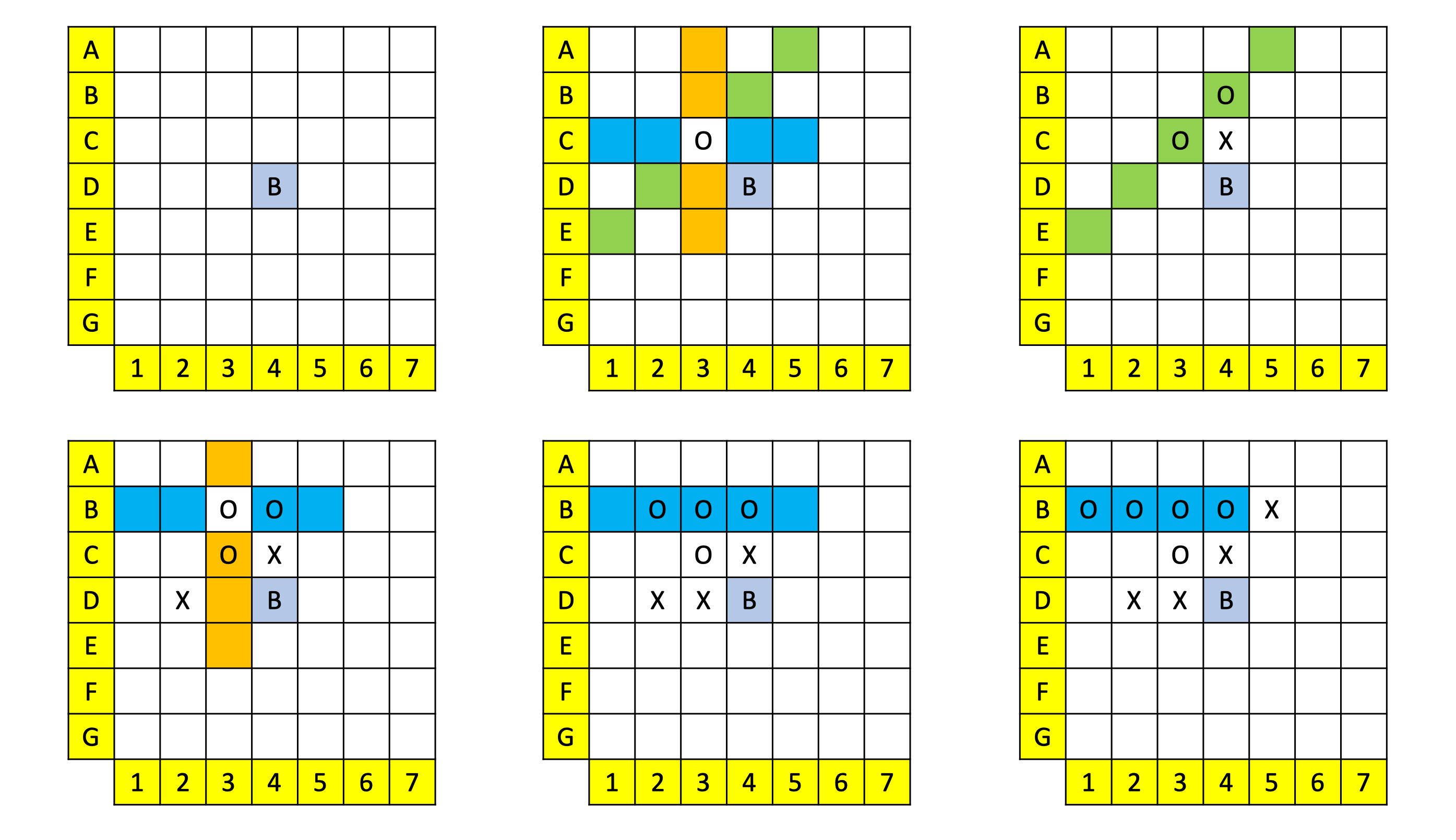}
		\caption{Winning sequence for Player 1 if the starting brick is at the centre (D3) location. Moves are read left to right, then top to bottom. First two board states show the starting brick position, and the first move by player 1 respectively. Thereafter, the next few board states show Player 2's (X) response followed by Player 1's (O) follow-up response.}
		\label{fig:variant1proof}
	\end{minipage}%
% 	\caption{add caption here}
\end{figure*}

\textbf{Illustration of Winning Strategy.} The winning strategy is shown in Fig. \ref{fig:variant1proof}. Initially, Player 1 should place his/her piece at C3, in order to form three threat axes (shown as the orange, blue and green axes). Player 2 can only block one of these axes in the next move as strongest play. Thereafter, Player 1 should expand to form a two-in-a-row in one of the unblocked axis (here shown as the green axis). Seeing as this can lead to an \textbf{Unblockable-3} the next turn, Player 2 will have to block this axis to prevent Player 1 from winning. Then, the third move for Player 1 will form a two-in-a-row with the first piece and the second piece (shown as the blue and orange axes), thereby ensuring his/her victory. Player 2 can only block one such axis, leaving Player 1 to form an \textbf{Unblockable-3} on the unblocked axis the next turn, and thereafter, winning the game.

\newpage
\subsection{Brick at E4 location (Variant 2)}

\begin{figure*}[h]
\centering
	\begin{minipage}[h]{\textwidth}
		\includegraphics[width=\textwidth]{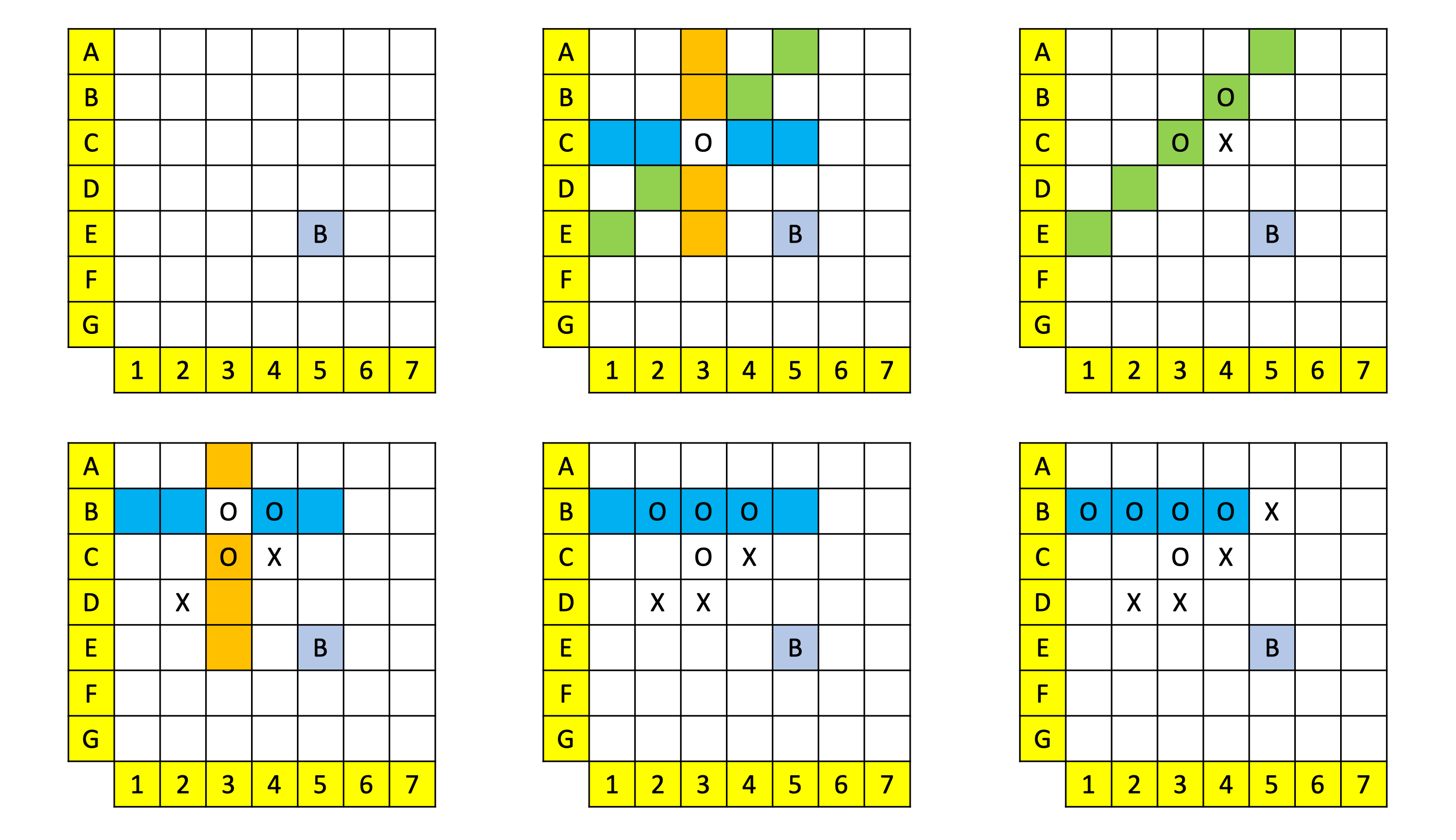}
		\caption{Winning sequence for Player 1 if the starting brick is at the E4 location. Moves are read left to right, then top to bottom. First two board states show the starting brick position, and the first move by player 1 respectively. Thereafter, the next few board states show Player 2's (X) response followed by Player 1's (O) follow-up response.}
		\label{fig:variant2proof}
	\end{minipage}%
% 	\caption{add caption here}
\end{figure*}

If the starting brick is not at the centre (D4), but instead at E5, then it is even easier for the first player to win as there is more open space on the top left corner of the board. One way for the first player to win is to apply the same set of moves as in Fig. \ref{fig:variant1proof}. This sequence of moves played out in Variant 2 is illustrated in Fig. \ref{fig:variant2proof}.

\newpage
\subsection{Brick at any location}

\begin{figure*}[h]
\centering
	\begin{minipage}[h]{\textwidth}
		\includegraphics[width=\textwidth]{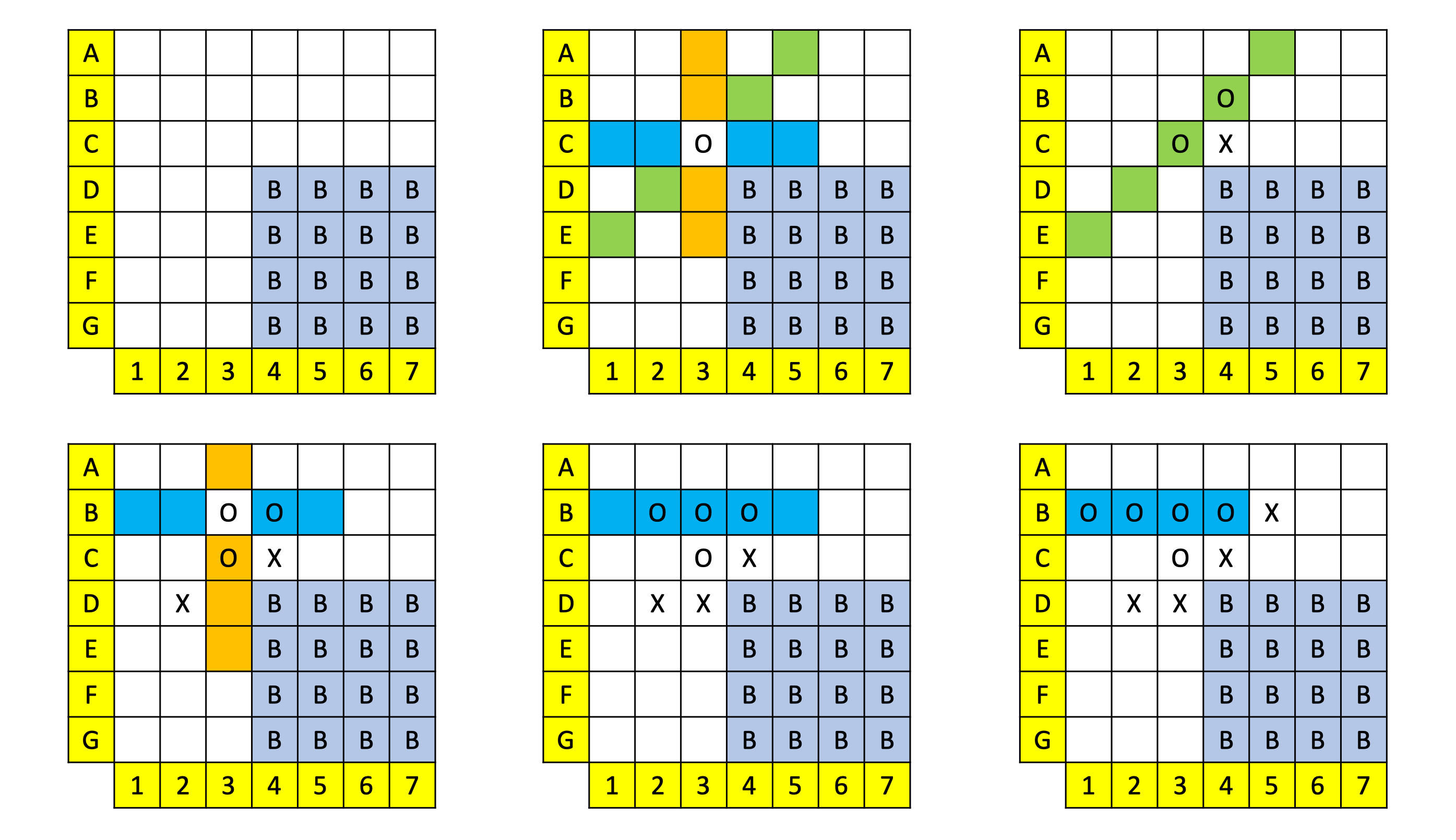}
		\caption{Winning sequence for Player 1 if the starting brick is at any location. Moves are read left to right, then top to bottom. First two board states show the starting brick position, and the first move by player 1 respectively. Thereafter, the next few board states show Player 2's (X) response followed by Player 1's (O) follow-up response.}
		\label{fig:allproof}
	\end{minipage}%
% 	\caption{add caption here}
\end{figure*}

In fact, we can generalize the argument made above to show that the brick can be at any starting position, and there will always be a winning sequence for Player 1. Without loss of generality, let us assume that the brick is always in the bottom right quadrant shown as the B locations in Fig. \ref{fig:allproof}. We can always assume this because BTTT is rotationally symmetric, and we can always rotate the grid such that the brick is at the bottom right quadrant. Thereafter, we apply the same argument for the brick in D4 or E5, and show that there exists a winning sequence for Player 1. In fact, for the brick at any position other than at D4, there will be more room for Player 1 to form threat axes and, in general, it should be easier for Player 1 to win.

\end{document}